\documentclass[journal]{IEEEtran}
\usepackage{amsmath,amsfonts}
\usepackage{algorithmic}
\usepackage{array}
\usepackage{textcomp}
\usepackage{stfloats}
\usepackage{url}
\usepackage{verbatim}
\usepackage{graphicx}

\usepackage{bm}
\usepackage{mathtools,cuted}
\usepackage{siunitx}
\usepackage{subcaption}
\usepackage{hyperref}
\usepackage{booktabs}

\usepackage[capitalize]{cleveref}
\crefname{equation}{}{}

\def\BibTeX{{\rm B\kern-.05em{\sc i\kern-.025em b}\kern-.08em
    T\kern-.1667em\lower.7ex\hbox{E}\kern-.125emX}}
\usepackage{balance}
\usepackage{lipsum}

\begin{document}

\title{Single-rod brachiation robot: Mechatronic control design and validation of pre-jump phases}

\author{
Juraj Lieskovsk\'{y}, Hijiri Akahane, Aoto Osawa, Jaroslav Bu\v{s}ek, Ikuo Mizuuchi, Tom\'{a}\v{s} Vyhl\'{\i}dal$^{\#}$% <-this % stops a space
\thanks{
    Received 24 January 2025; accepted 27 July 2025 in IEEE/ASME Transactions on Mechatronics, DOI: 10.1109/TMECH.2025.3598939; Dataset DOI 10.5281/zenodo.15576856
    
    This research was co-funded by the European Union under the project Robotics and Advanced Industrial Production  No.~CZ.02.01.01/00/22\_008/0004590, and by the Czech Science Foundation project 21--03689S. The first and the fourth author also acknowledge support by the Grant Agency of the Czech Technical University in Prague, student grant No.~SGS23/157/OHK2/3T/12.
}
\thanks{
    T. Vyhl\'{\i}dal ($^{\#}$corresponding author), J. Lieskovsk\'{y}, and Jaroslav Bu\v{s}ek are with the Dept.\ of Instrumentation and Control Eng., Faculty of Mechanical Engineering, Czech Technical University in Prague, Technick\'{a}~1902/4, 166~07 Praha~6, Czech Republic,
    {\tt Tomas.Vyhlidal@fs.cvut.cz}
}%
\thanks{
    Hijiri Akahane, Aoto Osawa and Ikuo Mizuuchi are with the Department of Mechanical Systems Engineering, Tokyo University of Agriculture and Technology, 2--24--16, Naka-cho, Koganei-City, Tokyo, Japan
}
\thanks{
    \textcopyright2025 The Authors. This work is licensed under a Creative Commons Attribution 4.0 License. For more information, see https://creativecommons.org/licenses/by/4.0/
}
}

%\markboth{Journal of \LaTeX\ Class Files,~Vol.~18, No.~9, September~2020}
%{How to Use the IEEEtran \LaTeX \ Templates}

\maketitle

%\blfootnote{\textcopyright2025 The Authors. This work is licensed under a Creative Commons Attribution 4.0 License. For more information, see https://creativecommons.org/licenses/by/4.0/}

\begin{abstract}
A complete mechatronic design of a minimal configuration brachiation robot is presented. The robot consists of a single rigid rod with gripper mechanisms attached to both ends. The grippers are used to hang the robot on a horizontal bar on which it swings or rotates. The motion is imposed by repositioning the robot's center of mass, which is performed using a crank-slide mechanism. Based on a non-linear model, an optimal control strategy is proposed, for repositioning the center of mass in a bang-bang manner. Consequently, utilizing the concept of input-output linearization, a continuous control strategy is proposed that takes into account the limited torque of the crank-slide mechanism and its geometry. An increased attention is paid to energy accumulation towards the subsequent jump stage of the brachiation. These two strategies are validated and compared in simulations. The continuous control strategy is then also implemented within a low-cost STM32-based control system, and both the swing and rotation stages of the brachiation motion are experimentally validated.
\end{abstract}


\section{Introduction}
Brachiation is a form of motion used by primates to move from one branch to another. Research into mimicking it with robots has previously been conducted mostly using multi-link mechanisms. The work by Fukuda et al. \cite{19901839}, where a six-link model of a brachiation robot was proposed and analyzed. In~\cite{291874} a two-link brachiating robot was developed, its motion being realized using heuristic control. A control system for this two-link robot was proposed in~\cite{843166} and torque time series minimizing the energy consumption of a two-link brachiation robot moving over a flexible cable was calculated in~\cite{8461036}.

\begin{figure}[tb!]
	\centering
    \includegraphics[width=.85\linewidth]{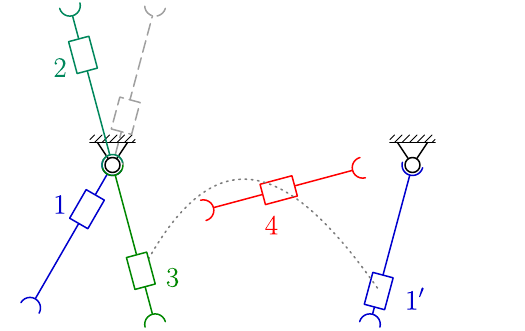}
    \caption{Phases of the robot's motion: 1 --- amplification of the swinging motion; 2 --- increasing of revolution speed; 3 --- preparation for release; 4 --- aerial phase}\label{fig:phases}
\end{figure}

In this paper, we focus on analysis, control optimization, and experimental validation of a minimal configuration of a brachiation robot. This extends the preliminary work presented by a part of the team in a conference paper~\cite{srbrachiation}, where a novel single-rod robot that uses an aerial phase in its motion was proposed and constructed. By periodically repositioning its center of mass while swinging or later rotating around a bar, the aim is to evoke sufficient energy to jump from one horizontal bar (branch) to another. The desired cycle of motion can be separated into four distinct phases; see \cref{fig:phases}. In the first two phases, energy is accumulated in the system, first during a swinging motion and secondly during rotation. The third phase is dedicated to preparation for the fourth phase, which is initiated by the robot releasing the bar, during which the robot spans the distance to the next bar. The phase ends by grasping the other horizontal bar and restarting the cycle. The general objective is to achieve locomotion of the robot between the bars. If the distances of bars are not equal, a different amount of energy must be determined and accumulated during the pre-jump phases for each distance. Note also that the distance between two neighboring bars must be greater than the length of the robot.

Compared to conventional brachiation robots with two-link arms with elbow joints, the advantage of the proposed single-rod robot is that it does not have the inherent chaotic dynamics of a serial multi-link pendulum. In terms of practical applications, it can be used as a means for moving through space on a pre-installed ladder-like structure. In the case of inspections of high-voltage lines, a single-rod brachiation robot can maintain its position without expending energy as long as it is holding a wire and is relatively robust to wind and external disturbances. Although the robot can only be used where there is a structure that can be grasped, it could be used for inspection work, maintenance work, and photography in spaces with overhead structures.

The preliminary work~\cite{srbrachiation} was focused on the first phase of the motion. Repositioning of the center of mass was performed by a feedforward policy, i.e. without any feedback from the angular position of the rod. Denoting the robot swing frequency as $\Omega$,  repositioning of the center of mass is to be performed with the frequency $2\Omega$. This proposed feedforward control is suitable for small swing angles, where the frequency $\Omega$ is (almost) constant. However, considering the physical pendulum-like nature of the robot, the oscillation frequency slightly reduces with growing amplitude of the swing. Thus, in the long run, this ideal to true frequency mismatch is likely to lead to loss of synchronization of the center-of-mass repositioning during the swing stage. Similarly, the synchronization loss can also be caused by disturbances acting on the robot, e.g. the effect of wind. In order to move the research closer to the applications, it was necessary to turn the feedforward control policy into a feedback policy, which can handle this synchronization imperfection through feedback from the swing angle measurement. The first attempts in this direction were presented in subsequent conference publication of the authors' team~\cite{ifac2023-brachiation}. Considering a simplified model of a rod with moving mass (free of the actuator dynamics), we proposed the feedback control of the first two phases of motion. It was done by utilizing the results on an analogous problem, a pendulum's swing and revolution. Such a \emph{swinging problem} occurs e.g.~in modeling children on a swing,~\cite{piccoli2005pumping, wirkus1998pump}. The oscillations of a pendulum with a periodically varying length were studied in~\cite{pinsky1999oscillations} with the key objective of determining the existence of periodic solutions. The stability of such periodic solutions was studied in~\cite{zevin2007qualitative, akulenko2009stability, belyakov2009dynamics}. The time-invariant control law to pump appropriate energy into the variable-length pendulum for achieving the desired swing motion was developed in~\cite{xin2014trajectory,xin2014variable}. A nonlinear feedback strategy to control the periodic motion of the pendulum has also been proposed in~\cite{reguera2016rotation} with a consideration of energy harvesting from the rotational motion.

An increased attention has also been paid to a related problem of using the pendulum length adjustment to dampen the pendulum swing. The open-loop solution derived in~\cite{STILLING200289} through energy analysis was turned by a part of the team into a practical closed-loop solution by introducing nonlinear time-delay feedback in~\cite{8003347}. The theoretical results are followed by comprehensive laboratory validation. In subsequent works,~\cite{8814293},~\cite{ECC2020} and~\cite{anderle2022controlling} the Lyapunov method was applied to derive the nonlinear control rule to damp the pendulum swing.
In~\cite{li2019improved}, the efficiency of amplitude suppression of an oscillating pendulum by a controllable moving mass was studied by simulations for several suppression rules. The problem was further studied, and its results were experimentally validated in~\cite{li2020experimental}. Let us also point to an analogous problem by the author's team studied in~\cite{kuvre2021algorithms}, where the pendulum length is kept fixed, and its angular motion is damped by an up-and-down motion of the pivot. 

In~\cite{anderle2022controlling}, in addition to the design and analysis of Lyapunov-based control rules, a numerical study was performed to determine the optimal solution. For the damping of the pendulum, it leads to a bang-bang length variation, where the pendulum is stepwise prolonged when passing the (equilibrium) zero-angle position and stepwise shortened at the turning-angle positions. Although this damping problem is inverse to the first phase of the motion of the brachiation robot considered here, the results can be applied well if the direction of motion of the center of mass is reversed. This idea was applied and elaborated further in~\cite{ifac2023-brachiation} for the first stage of the brachiation motion, and was also adapted for the second stage of motion.

Beyond the preliminary results presented in the conference papers~\cite{srbrachiation} and~\cite{ifac2023-brachiation}, and beyond the state of the art, the contribution of this paper is as follows:
\begin{IEEEitemize}
\item Compared to~\cite{srbrachiation}, the construction of the brachiation robot is adjusted to allow model-based validation of the swing and rotation stages of the brachiation motion. 
In addition, the control system and measurement hardware are redesigned to allow experimental validation of feedback control in the swing and rotation stages.
\item The feedback control policies proposed conceptually in~\cite{ifac2023-brachiation} for a mathematical model of a rod with moving mass are adapted for the experimental setup of the brachiation robot. 
\item For the objective of the control design, a precise mathematical model of the robot is derived and parameterized, including the submodel of the crank-slide mechanism used to move the center of mass (not considered in~\cite{ifac2023-brachiation}).  
\item Next to the validation of the control design concepts in the mathematical model, for the first time, the experimental validation of the swing and rotation stages of the brachiation by a single-rod robot is performed.     
\end{IEEEitemize}

The remainder of the paper is structured as follows. In \cref{sec:problem}, utilizing the results of~\cite{ifac2023-brachiation}, we define the optimal motion of the center of mass and the two targeted phases of single-rod robot's brachiation. 
In \cref{sec:experiment} we present the experimental setup for its validation, including its construction and hardware adjustments compared to \cite{srbrachiation}. In \cref{sec:model} a precise nonlinear model of the setup is derived and parameterized. The ideal {\em limit case} and practically applicable {\em continuous control} policies are proposed in \cref{sec:policies} using the robot's model. In \cref{sec:validation} a thorough case study validation is performed. It includes a simulation-based validation of both control policies, followed by an experimental validation of the continuous control policy. The main results and research prospects are provided in the concluding \cref{sec:conclusions}.

\section{Optimal motion of the center of mass}\label{sec:problem}

In the conceptual work~\cite{ifac2023-brachiation}, it was shown that a time-optimal policy for the amplification of the swing of the single-rod brachiation robot calls for the stepwise repositioning of its center of mass, mirroring the approach described in~\cite{anderle2022controlling} for the optimal swing damping of a variable length pendulum. It was also shown that stepwise repositioning is time-optimal in the second phase of the robot's motion when it revolves around the bar, requiring a minor adjustment in the control algorithm. The optimal policy can be best analyzed when viewed through the two mechanisms by which the total energy of the system can change.

The first mechanism is the change in potential energy. Everything else being stationary, moving the mass towards or away from its axis of rotation either increases or decreases the potential energy of the system. The maximum amount of potential energy can then be added to the system by moving the mass from one limit of its relative position to the other in either of the robot's equilibriums. This also holds when performed instantaneously while the robot's body rotates, giving us the optimal limit-case policy from the perspective of potential energy for phase two. During the first phase of motion, where the unstable equilibrium is never reached, we may then minimize the loss or maximize the gain in potential energy (depending on whether the robot is beyond the horizontal plane) by extending the mass away from the axis of rotation at the turning angle, completing the periodic motion.

\begin{figure}
	\begin{subfigure}{0.48\linewidth}
		\includegraphics[scale=0.5]{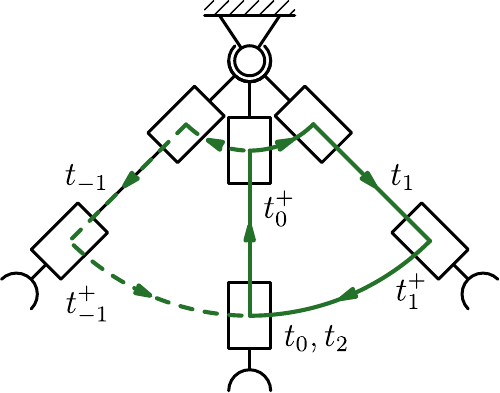}
		\subcaption{Full period of control during the first (swinging) phase}
	\end{subfigure}
	\hspace*{\fill}
	\begin{subfigure}{0.48\linewidth}
        \centering
		\includegraphics[scale=0.5]{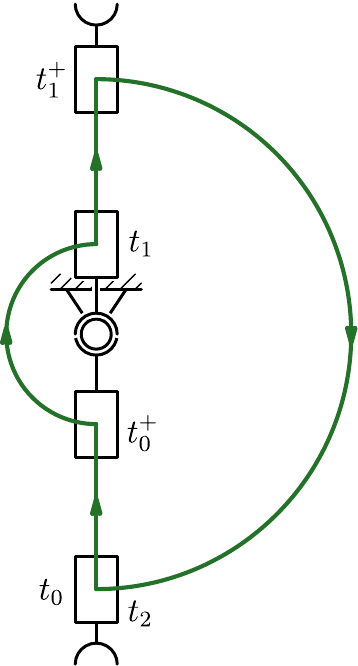}
		\caption{Full period of control during the second (rotation) phase}
	\end{subfigure}
	\caption{The general control policy for the first two phases of the robot's motion. % \cite{ifac2023-brachiation}, see video of a simulation\protect\footnotemark.
    }\label{fig:policies}
\end{figure}

The second mechanism is the effect of the Coriolis force, caused by the relative motion of the mass, on the system's kinetic energy. In the broadest terms, the Coriolis force aids in angular motion when the mass is retracted towards the robot's axis of rotation and acts against it when extended away from the axis of rotation. As it is proportional to the body's angular velocity, it follows that, for the first phase, the mass should be retracted when passing the stable equilibrium and extended at the turning angle. Similarly, during the second phase, the weight should be retracted at the same position and then extended at the unstable equilibrium, where the angular velocity is the lowest.

As we can see, both mechanisms prescribe the same general motion, giving us a limit-case policy for the first two phases, which asks for the mass to be repositioned to the opposite extreme after each half of a swing/rotation, as shown in \cref{fig:policies}, where its position immediately before and after each event is marked in sequence by $t_i$ and $t_i^+$, respectively.

\section{Experimental Setup}\label{sec:experiment}

\begin{figure}[tb!]
    \centering
    \begin{subfigure}{\linewidth}
        \includegraphics[width=\textwidth]{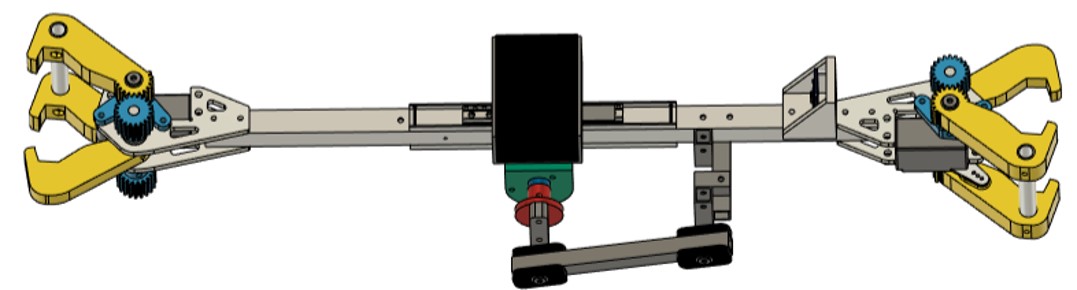}
        \caption{{CAD} design model of the single-rod robot}
    \end{subfigure}
    \\
    \begin{subfigure}{\linewidth}
        \includegraphics[width=\textwidth]{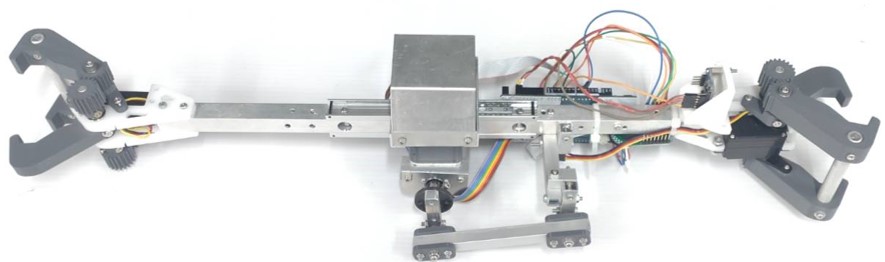}
        \caption{Experimental setup single-rod robot}
    \end{subfigure}
    \\
    \begin{subfigure}{0.55\linewidth}
        \includegraphics[width=\textwidth]{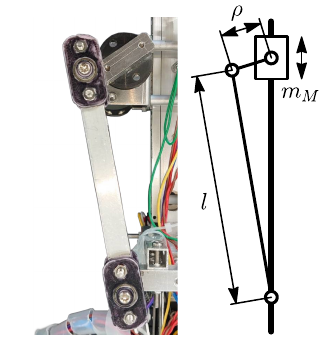}
        \caption{Detail of the crank-slide mechanism}
    \end{subfigure}
    \begin{subfigure}{0.38\linewidth}
        \includegraphics[width=\textwidth]{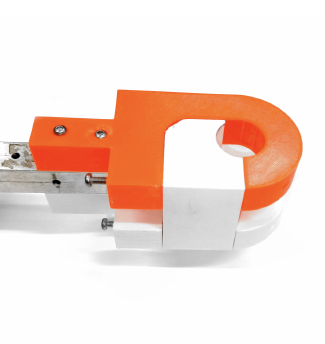}
        \caption{Simplified gripper used for the experiment}
        \label{fig:grip-detail}
    \end{subfigure}
    \caption{Mechanical design and main components of the single-rod brachiation robot}
    \label{fig:3_1}
\end{figure}

The experimental setup shown in \cref{fig:3_1} consists of a rigid square aluminum rod of length $\SI{610}{\milli\meter}$ with two identical 3D printed gripper mechanisms, actuated by servo modules (Hitec HS8775MG), attached to both ends. The servos allow the grippers to grab and release a round bar during the brachiation. For the purpose of experiment repeatability in validating the controlled swing and rotation phases of the motion, one of the grippers was replaced by a clamp-like simplified gripper allowing a rigid connection with the horizontal bar, nested in bearings, see \cref{fig:grip-detail}. A crank-slide mechanism placed on the setup transforms the rotational motion of a {BLDC} (Brushless Direct Current) motor Maxon EC22 40W with a reduction ratio of 1:128 into linear motion of a $\SI{0.886}{\kilo\gram}$ weight along the main bar, with a $\pm\SI{20}{\milli\meter}$ range of motion. The entire experimental setup weighs $\SI{1.473}{\kilogram}$.

\begin{figure}[tb!]
    \centering
    \includegraphics[width=1\linewidth]{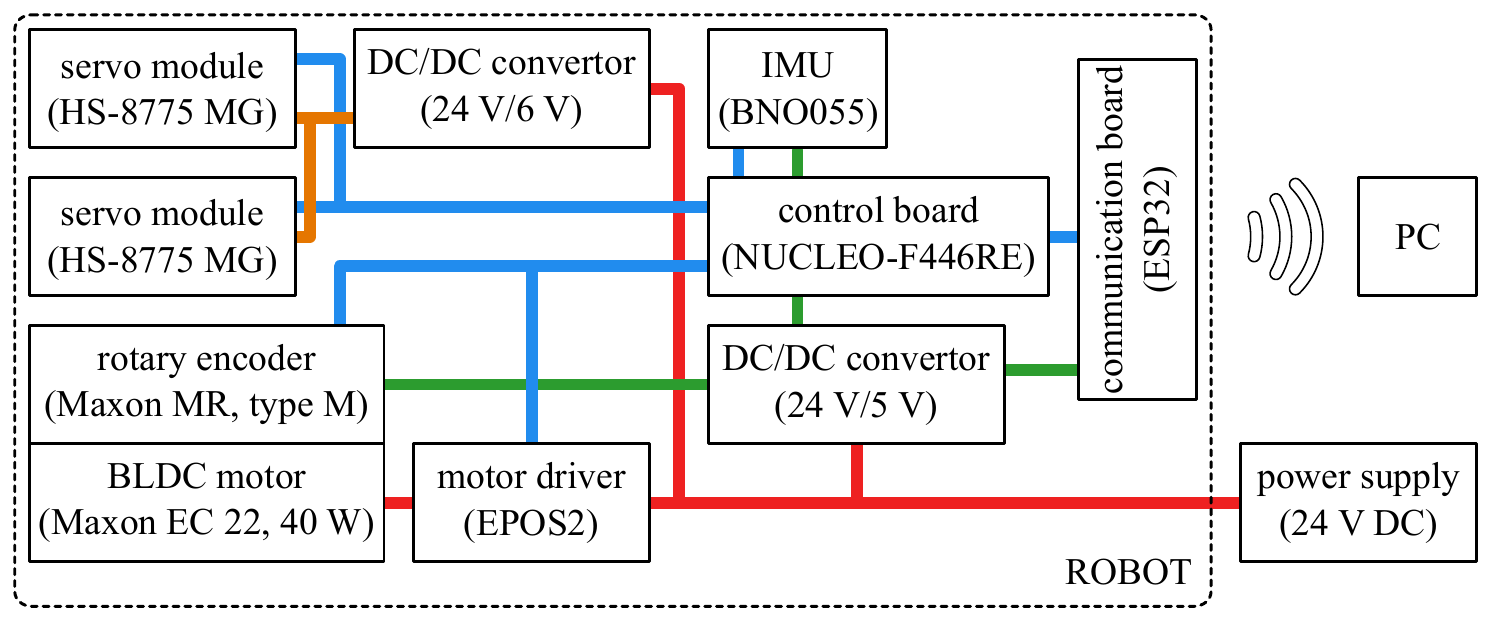}
    \caption{Block diagram of the experimental setup hardware components.}
    \label{fig:3_3}
\end{figure}

Concerning the control circuit of the robot, it is shown in \cref{fig:3_3}. The experimental setup is controlled by a STM32 NUCLEO-F446RE board mounted on the robot's body. The NUCLEO board is connected to an ESP32 board using an UART. The ESP32 board transfers data wirelessly to a PC over a Wi-Fi connection. The motor of the crank-slide mechanism is driven by Maxon EPOS2 with a Maxon MR type M rotary encoder as feedback. The attitude and angular velocity of the experimental setup are measured by a Bosch BNO055 {IMU} (Inertial Measurement Unit) thanks to Bosch Sensortec sensor fusion software included onboard. A 24V DC power supply is used to power the setup in combination with two DC/DC step-down converters (\SI{6}{\volt} for the servos and \SI{5}{\volt} for the control board and sensors). Note also that compared to~\cite{srbrachiation}, the originally used brushed motor was replaced by a BLDC motor including a current control driver. In addition, the sensor used to detect the position of the weight was changed from a linear potentiometer to a rotary encoder, so that the angle of the motor moving the weight could be detected directly, rather than inferred from the position of the weight.

\section{Dynamical Model of Single-Rod Robot}\label{sec:model}
\begin{figure}[tb!]
	\centering
	\includegraphics[width=\linewidth]{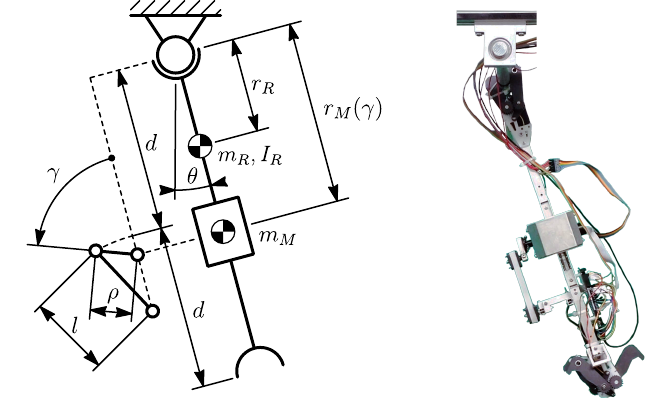}
	\caption{A scheme of the single-rod brachiation robot (crank-slide mechanism in side-view)}\label{fig:scheme}
\end{figure}

The body of the robot in the experimental setup, described in \cref{sec:experiment}, is modeled as a rod (rigid body), rotating around the origin of an inertial frame, and a point mass that translates along its dominant dimension, perpendicular to the axis of rotation. The scheme of the simplified structure is shown in \cref{fig:scheme}. The rod is characterized by its mass $m_R$, the distance of its center of mass from the axis of rotation $r_R$, and the moment of inertia about its center of mass $I_R$. The point mass is then described by its relative position $r_M$, given by the geometry of the crank-slide mechanism, and its mass $m_M$. The inertial properties of the crank-slide mechanism are simplified to a single parameter $I_S$ that represents the combined inertia of the crank, gearbox, and motor. As a whole, the system has two degrees of freedom, reflected in the number of generalized coordinates used to describe its dynamics. Those were chosen as the angle $\theta$ by which the rod deviates from the vertical axis and the angle of the crank-slide mechanism $\gamma$. The relative position $r_M$ is then a function of the generalized coordinate $\gamma$, radius of the crank $\rho$, length of the connecting rod~$l$ and the mean position~$d$.

% \section{Equations of motion}\label{sec:eom}

For a vector of generalized coordinates $\bm{q} = \begin{bmatrix} \theta & \gamma \end{bmatrix}^\top$, the system's equations of motion are derived by expanding the terms of Lagrange's equations of the second kind, namely by substituting the kinetic $T(\bm{q}, \dot{\bm{q}})$ and potential $V(\bm{q})$ energy individually for the system's Lagrangian $L = T - V$ and applying the chain rule when taking derivatives with respect to time, resulting in the form
\begin{equation}\label{eqn:EoM}
	\underbrace{
	\frac{\partial^2 T}{\partial \bm{\dot{q}}^2}
	}_{\bm{M(\bm q)}} \bm{\ddot{q}}
	+
	\underbrace{
	\frac{\partial^2 T}{\partial \bm{\dot{q}} \partial \bm{q}} \bm{\dot{q}} - \frac{\partial T}{\partial \bm{q}}
	}_{\bm{c}(\bm{q},\bm{\dot{q}})}
	+
	\underbrace{
	\frac{\partial V}{\partial \bm{q}}
	}_{-\bm{\tau}_p(\bm q)}
	=
	\bm{\tau}(\bm q, \bm{\dot{q}}),
\end{equation}
where $\bm{M}(\bm{q})$ is the mass matrix, $\bm{c}(\bm{q},\bm{\dot{q}})$ is a vector of torques resulting from Coriolis and centrifugal forces acting on the system, $\bm{\tau}_p(\bm{q})$ is a vector of torques tied to the potential energy of the system and $\bm{\tau}(\bm q, \bm{\dot{q}})$ is a vector of external torques. The vector $\bm{\tau}(\bm q, \bm{\dot{q}})$ can be split into two terms:
\begin{equation}\label{eqn:EoM-torques}
	\bm{\tau}(\bm q, \bm{\dot{q}}) = \bm{d}(\bm q, \bm{\dot{q}}) + \bm{B} \bm{u},
\end{equation}
where $\bm{d}(\bm q, \bm{\dot{q}})$ is a vector of torques caused by viscous damping and the matrix $\bm{B}$ redistributes the control input $\bm{u}$ containing a single element which corresponds to the torque exerted by the {BLDC} motor of the crank-slide mechanism. The model of the system given by \cref{eqn:EoM} and \cref{eqn:EoM-torques} can be written in the compact form
\begin{equation}
\label{eqn:Model}
	 \bm{M(\bm q)} \bm{\ddot{q}}+\bm{c}(\bm{q},\bm{\dot{q}})-\bm{\tau}_p(\bm q)= \bm{d}(\bm q, \bm{\dot{q}}) + \bm{B} \bm{u}.
\end{equation}

%The generalized coordinates are chosen as $\bm{q} = \begin{bmatrix} \theta & \gamma \end{bmatrix}^\top$ with an explanation of the decision provided later in \cref{sec:continuous}. For these coordinates, 
Considering the system structure as shown in \cref{fig:scheme}, its kinetic and potential energy are given as
\begin{subequations}
\begin{align}
	T(\bm{q},\dot{\bm{q}}) &= 
	\frac{1}{2} (
	m_R \, r_R^2 \, \dot{\theta}^2 + I_R \, \dot{\theta}^2 +
	m_M \, r_M^2{(\gamma)} \, \dot{\theta}^2 \notag \\
	&\phantom{=}+ m_M \, \dot{r}_M^2{(\gamma,\dot{\gamma})} + I_S \, \dot{\gamma}^2 ),
	\label{eqn:T}
	\\
	V(\bm{q}) &= m_R \, g \, (r_R-r_R \, \cos(\theta)) \notag \\
        &\phantom{=}+ m_M \, g \, (r_M(0) - r_M(\gamma) \, \cos(\theta)), \label{eqn:V}
\end{align}
\end{subequations}
where
\begin{align*}
	r_M(\gamma) &= d + \rho \, \cos(\gamma) + e(\gamma), \\ 
	e(\gamma) &= \pm \left(l - \sqrt{l^2 - \rho^2 \sin^2(\gamma)}\right).
\end{align*}
The sign in the term $e(\gamma)$ depends on which gripper is grasping the bar (``$+$'' for the configuration in \cref{fig:scheme} and ``$-$'' in the experiment). Furthermore, assuming $l\gg\rho$, the term can be discarded without a significant loss in accuracy.
%We do so in the on-board implementation of the proposed control policy as in our opinion calculating partial derivatives of $e$ is in our case not worth the computational load, considering other model uncertainties.}
Consequently, application of a computer algebra system ({CAS}) yields terms on the left-hand side of \cref{eqn:Model}:
\begin{align*}
    \bm{M}(\bm{q})
    &=
	\scalebox{0.8}{$
    \begin{bmatrix}
        I_R + m_M r_M^2(\gamma) + m_R r_R^2 & 0 \\ 0 &  I_S + m_M {\left(\frac{\partial r_M}{\partial \gamma}(\gamma)\right)}^2
    \end{bmatrix},
	$}
    \\
    \bm{c}(\bm{q},\dot{\bm{q}})
    &=
	\scalebox{0.8}{$
    \begin{bmatrix}
        2 m_M r_M(\gamma) \frac{\partial r_M}{\partial \gamma}(\gamma) \, \dot{\gamma} \, \dot{\theta}\\
        m_M\left(\frac{\partial r_M}{\partial \gamma}(\gamma)\frac{\partial^2 r_M}{\partial \gamma^2}(\gamma) \, \dot{\gamma}^2 - r_M(\gamma)\frac{\partial r_M}{\partial \gamma}(\gamma)\,\dot{\theta}^2\right) 
    \end{bmatrix},
	$}
    \\
    \bm{\tau}_p(\bm{q})
    &=
	\scalebox{0.8}{$
    \begin{bmatrix}
        -g \sin{(\theta)} \left(m_M r_M(\gamma) + m_R r_R\right) \\
        g m_M \frac{\partial r_M}{\partial \gamma}(\gamma) \cos{(\theta)}
    \end{bmatrix}.
	$}
\end{align*}

Concerning external forces acting on the model, a viscous damping model is chosen to model passive forces in each bearing. The effects of the rotary bearings and the input act in the direction of the generalized coordinate, while the friction of the linear guide of the crank-slide mechanism has to be transformed into generalized coordinates by pre-multiplying the force in physical coordinates $\bm{F}$ by the transpose of $\frac{\partial \bm{r}}{\partial \bm{q}}$, where $\bm{r}$ is a vector towards its point of effect. Therefore, the terms that complete the right-hand side of \cref{eqn:Model} are given as
\begin{align*}
	\bm{d}(\bm{q},\dot{\bm{q}})
	&=
	\begin{bmatrix}
	-b_R \, \dot{\theta} \\
	-b_C \, \dot{\gamma} - b_S\,\left(\frac{\partial r_M}{\partial \gamma}(\gamma)\right)^2\dot{\gamma}
	\end{bmatrix},
	\\
	\bm{B}
	&=
	\begin{bmatrix}
	0 \\ 1
	\end{bmatrix},
\end{align*}
where $b_{\mathrm{R}}$, $b_{\mathrm{C}}$, and $b_{\mathrm{S}}$ are damping coefficients in the bearings of the rod, crank, and slide, respectively.

\subsection{State-space Description}

The state-space description of the system was formed for a state vector defined as $\bm{x} = [\bm{q}^\top \ \bm{v}^\top]^\top$, where $\bm{v} = \dot{\bm{q}}$. The dynamics of the system are then given by
\begin{equation}\label{eqn:state-space}
    \scalebox{0.9}{$
    \dot{\bm{x}}
	=
	\begin{bmatrix}
		\bm{v} \\	\bm{M}^{-1}(\bm{q}) \left( - \bm{c}(\bm{q},\bm{v}) + \bm{\tau}_p(\bm{q}) + \bm{d}(\bm{q},\bm{v})+\bm{B}\bm{u}\right)
	\end{bmatrix}.
    $}
\end{equation}

This representation of the system is used to perform simulations in \cref{sec:validation:lc} and \cref{sec:validation:pd}. Additionally, to formulate a control policy in \cref{sec:policies:limitcase} 
the system's generalized momentum $\frac{\partial L}{\partial \bm{\dot{q}}}$ was formalized in the convenient matrix form
\begin{equation}\label{eqn:momentum}
	\bm{p}(\bm{x}) = \bm{M}(\bm{q})\,\bm{v}.
\end{equation}

\section{Control Policies}\label{sec:policies}

Having formulated the system's state-space representation, we may now describe two control policies following the principles outlined in \cref{sec:problem}. First, a limit case where the weight can be repositioned instantaneously, and second, its approximation, feasible with limited torques.

\subsection{Limit Case Control Policy}\label{sec:policies:limitcase}

In this limit case, instantaneous repositioning allows us to directly quantify the amount of mechanical energy added or subtracted from the system based on the state preceding the shift. In order to do so, the system must be analyzed in the context of hybrid dynamics~\cite{Goebel2009}, which allows the system to exhibit both continuous and discrete behavior. This leads to the following model
\begin{align}
	\dot{\bm{x}} &= F(\bm{x}), \quad \bm{x} \in C, \\
	\bm{x}^+ &= G(\bm{x}), \quad \bm{x} \in D,
 \label{eqn:switch}
\end{align}
where $G: \mathbb{R}^n \rightarrow \mathbb{R}^n$ is a jump map, which instantaneously changes the current state of the system $\bm{x} \in \mathbb{R}^n$ to a new state $\bm{x}^+ \! \in \mathbb{R}^n$, $D \subset \mathbb{R}^n$ is a jump set of states the occurrence of which triggers a jump described by the jump map, $C = \mathbb{R}^n \setminus D$ is a flow set and $F: \mathbb{R}^n \rightarrow \mathbb{R}^n$ is a differential equation capturing the continuous dynamics of the system.

To reason about changes of the system's state caused by jumps we will utilize impulses (of force), defined for an external force acting on the system during an infinitesimally short time interval, as
\begin{equation*}
	\bm{J} = \lim_{t \rightarrow t^+} \int_{t}^{t^+} \bm{F} \, \mathrm{d}t = \bm{p}(\bm{x}^+) - \bm{p}(\bm{x}) \ ,
\end{equation*}
where times $t$, $t^+$ coincide with states $\bm{x}$ and $\bm{x}^+$, $\bm{F}$ is the external force and $\bm{p}$ the momentum of the system. With some limitations, integration of the system's dynamics can be bypassed by directly evaluating its momentum. Instantaneous repositioning of the weight can then be explained as the system's reaction to an impulse of force, with both positive and negative values during its infinitesimal duration, delivered by an ideal actuator.

A jump set, which contains the states of the system's representation in \cref{eqn:state-space} at which the weight should be repositioned as outlined in \cref{sec:problem}, can be defined as a union of three sets $D = \cup_{i=1}^3 D_i$
\begin{subequations}\label{eqn:jump_set:all}
    \begin{align}
    	D_1 &= \left\{ \bm{x} \in \mathbb{R}^4: \cos(\theta) = 1 \,,\; \dot{\theta} \neq 0 \right\} \label{eqn:jump_set:D1},\\
    	D_2 &= \left\{ \bm{x} \in \mathbb{R}^4: \cos(\theta) \neq 1 \,,\; \dot{\theta} = 0 \right\}, \label{eqn:jump_set:D2}\\
    	D_3 &= \left\{ \bm{x} \in \mathbb{R}^4: \cos(\theta) = -1 \,,\; \dot{\theta} \neq 0 \right\} \ , \label{eqn:jump_set:D3}
    \end{align}
\end{subequations}
where $D_1$ and $D_3$ correspond to the robot's body passing the stable or unstable equilibrium and $D_2$ the turning angle being reached. 
The jump map returns states where $\gamma \in \{0,\pi\}$ and $\dot{\gamma} = 0$, depending on whether the weight should be moved toward or away from the robot's axis of rotation. To determine the values of $\theta$ and $\dot{\theta}$ we analyze the effects of an impulse delivered by the motor on the robot's momentum expressed in generalized coordinates; see \cref{eqn:momentum}. As the impulse acts only in the direction of the second coordinate, the position and momentum in the first coordinate have to be conserved. This gives us equations
\begin{align}
	\theta^+ &= \theta, \label{eqn:conservation:position}\\
	M_{11}(\gamma^+)\,\dot{\theta}^+ &= M_{11}(\gamma)\,\dot{\theta},\label{eqn:conservation:velocity} 
\end{align}
which determine the values of $\theta^+$ and $\dot{\theta}^+$ in the jump map
\begin{equation} \label{eqn:jump_map}
	G(\bm{x})
	=
	\begin{cases}
		\begin{bmatrix} \theta & \pi  & \frac{M_{11}(\gamma)}{M_{11}(\pi)}\dot{\theta} & 0 \end{bmatrix}^\top, & \bm{x} \in D_1, \\[5pt]
		\begin{bmatrix} \theta & 0 & \frac{M_{11}(\gamma)}{M_{11}(0)}\dot{\theta} & 0 \end{bmatrix}^\top, & \bm{x} \in D_2 \cup D_3.
	\end{cases}
\end{equation}

Having defined the jump set and the jump map, we may redirect our focus to the dynamics within the flow set $C = \mathbb{R}^4 \setminus D$. The approach asks for maintaining the crank-slide mechanism in the state $\gamma^+ \in \{0, \pi\}$, $\dot{\gamma}^+ = 0$ set by the jump map, which can be achieved by prescribing the input as
\begin{equation*}
	u = -d_2(\gamma,\dot{\gamma}) + c_2(\gamma,\dot{\theta},\dot{\gamma})-\tau_{p_2}(\theta, \gamma).
\end{equation*}
Based on \cref{eqn:conservation:position} and \cref{eqn:conservation:velocity} the change in the kinetic and potential energy of the system associated with each state jump can be derived as
\begin{subequations}  \label{eqn:ene_change}
    \begin{align}
	T(\bm{x}^+) - T(\bm{x}) &= \frac{1}{2} M_{11}(\gamma) \left(\frac{M_{11}(\gamma)}{M_{11}(\gamma^+)} - 1\right) \dot{\theta}^2, \label{eqn:kin_change}\\
	V(\bm{x}^+) - V(\bm{x}) &= - m_M \, g \, (r_M(\gamma^+) - r_M(\gamma)) \cos(\theta) \label{eqn:pot_change}\,.
\end{align}
\end{subequations}
For jumps defined by the map in \cref{eqn:jump_map} the potential energy change in \cref{eqn:pot_change} further simplifies to
\[
	V(\bm{x}^+) - V(\bm{x}) = \pm \, 2 \, m_M \, g \, \rho \, \cos(\theta).
\]
 
\subsection{Continuous Control Policy}

The continuous control policy transforms the previously described limit case policy into one that is practically applicable by considering actuation limits. It results from the analysis of the optimal solution in \cref{sec:problem} that asks for the angle $\gamma$ to be changed between the extreme values $\gamma = \{0,\pi\}$ as fast as possible. Taking into account the two phases visualized in \cref{fig:policies}, the map of the switching points described in \cref{eqn:jump_set:all} can be projected to the desired value of $\gamma$ denoted as its setpoint $\gamma_d$, given as 
\begin{equation}
\label{eqn:gamma_d}
	\gamma_\mathrm{d}(\theta, \dot{\theta}) = \frac{\pi}{2} (1 + \operatorname{sign}(\sin(\theta)) \operatorname{sign}(\dot{\theta})),
\end{equation}
which depends purely on the motion of the robot's body, i.e.\ on its angle $\theta$ and angular velocity $\dot\theta$.

The control design objective is to synthesize such a control input $u$, which turns the governing equation for the crank-slide angle
\begin{equation} \label{eqn:ddgamma}
	\ddot{\gamma} = M_{22}^{-1}(\gamma) (u + d_2(\gamma,\dot{\gamma}) -c_2(\gamma,\dot{\theta},\dot{\gamma})+\tau_{p_2}(\theta, \gamma)),
\end{equation}
taken from \cref{eqn:Model}, into a linear second-order system
\begin{equation}
	\ddot{\gamma} + 2 \zeta \omega \dot{\gamma} + \omega^2\gamma = \omega^2\gamma_\mathrm{d}(\theta,\dot{\theta}),
 \label{eqn:2ndorder}
\end{equation}
with a natural frequency $\omega$ and damping coefficient $\zeta$, both of which can be tuned. As the first step, we introduce a control signal
\begin{equation}\label{eqn:signal}
	w = - \omega^2 (\gamma - \gamma_\mathrm{d}(\theta,\dot{\theta})) - 2 \zeta\omega \dot{\gamma}.
\end{equation}
Then, utilizing the concept of input-output linearization, we form the control input as
\begin{equation}\label{eqn:input}
	u = M_{22}(\gamma) w - d_2(\gamma,\dot{\gamma}) + c_2(\gamma,\dot{\theta},\dot{\gamma}) - \tau_{p_2}(\theta, \gamma).
\end{equation}
Combining \cref{eqn:input} and \cref{eqn:signal} and substituting the result into \cref{eqn:ddgamma} then produces the desired form \cref{eqn:2ndorder}. 

Concerning the parameterization of the desired dynamics, the choice $\zeta = 1$ is reasonable as it provides a critically damped response. The speed of the response is then determined by the frequency $\omega$. From a performance point of view, the higher $\omega$, the better. However, the torque limits of the motor must be taken into account. The peak torque requirement occurs when the setpoint changes after the crank-slide mechanism has reached a steady state in either the top or bottom position. At that instant $\gamma - \gamma_\mathrm{d} \approx \pm \pi$, $\dot{\gamma} = 0$ and $M_{22} = I_S$, which corresponds to the absolute value of the desired input
\begin{equation} 
	|u| \approx I_S \omega^2 \pi \ .
 \label{eqn:omega}
\end{equation}
Substituting $|u|$ by the maximum torque that can be provided by the motor, we may express an estimate of the largest possible value of $\omega$ that satisfies our requirement.

An additional point which we aim to discuss is related to the system non-controllability at its stable equilibrium, that is, for the initial condition $\theta(0)=0$ and $\dot\theta(0)=0$. In this singular position, moving the mass $m_M$ does not project to a variation in momentum needed to initiate the swing motion. Note that for its initiation, such a deflection of the pendulum is needed, under which the induced momentum overcomes the effect of friction in the holding gripper. Naturally, also such an actuator and mechanism are needed which provides considerably higher injected energy compared to the energy dissipated due to friction; see~\cite{califano2022energy} studying an interplay between energy injection and dissipation over a motion cycle.

To conclude, note that the above analysis can also be utilized for optimizing the robot's structure. The amount of injected energy is positively related to the range of the crank-slide mechanism, the mass $m_M$, and its velocity. Therefore, the ratio between the injected and the total energy required for the aerial maneuver can be maximized to enhance performance. For example, since the total required energy is proportional to the total mass of the robot, it is desirable to concentrate it as much as possible into the moving mass $m_M$.

\section{Case study validation}\label{sec:validation}

\begin{figure}[tb!]
    \centering
    \includegraphics[width=\linewidth]{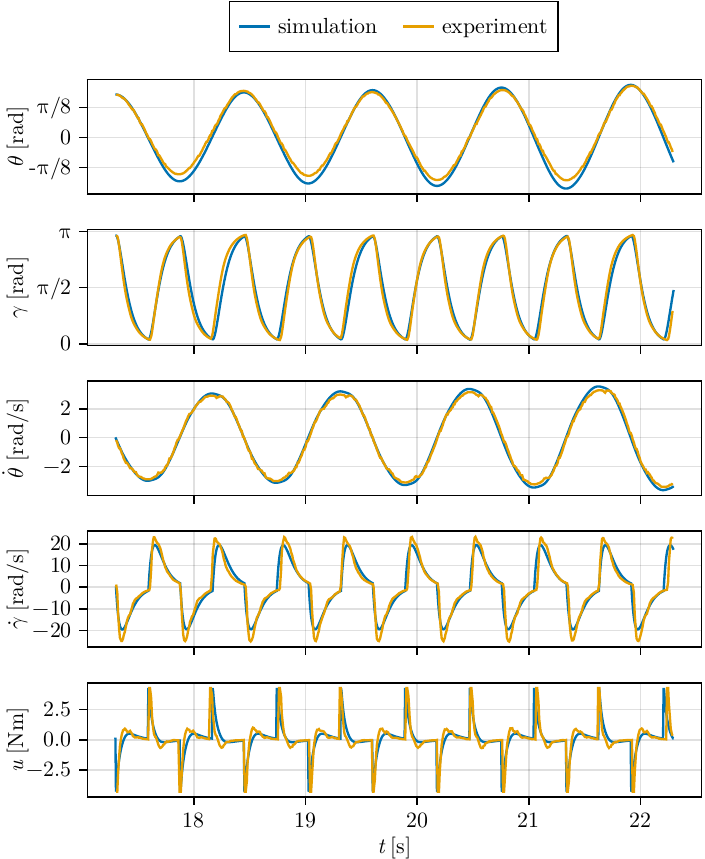}
    \caption{Comparison of closed-loop simulation and experimental  results. The initial state of the simulation is identical with the estimated at the start of the captured segment.}\label{fig:comparison}
\end{figure}

First, both control policies described above are tested in simulations. They are performed using the state-space model \cref{eqn:state-space} with the parameters given in \cref{tab:parameters}. The model's parameters were identified offline in two phases using joint moving horizon estimation~\cite{michalskaMovingHorizonObservers1995} (both state and parameter) and an additional smoothing step in order to attain constant parameter estimates. Except for $\rho$, $d$, and the total mass of the robot that were measured directly, all parameters of the system (in \cref{tab:parameters}), were identified in the first phase based on a closed-loop experiment with a more basic PID controller (parameterized heuristically). In the second phase, estimates of parameters $I_R$, $r_R$, and $b_R$, which do not appear in the proposed controller ($m_R$ is tied to $m_M$ through the robot's total mass), were re-identified to particularly fit a short segment ($\SI{17.3}{\second} < t < \SI{22.3}{\second}$) of the final experiment shown in \cref{fig:experiment:states}. To illustrate the result of the identification process, closed-loop simulated response of the identified model is compared to the experiment in \cref{fig:comparison}. As can be seen, a very good system-model match has been achieved for the identified parameter set. Note that the slight differences seen mainly in the actuation are probably caused by neglecting dry friction. However, notice that the amplitude and phase match of the periodic motions are close to being ideal.

\begin{table}[tb!]
	\centering
	\begin{tabular}{crl}
		Parameter & Value & Unit \\ \midrule
		$m_R$ & 0.587 & \si{\kilogram} \\
		$I_R$ & 2.64e-2 & \si{\kilogram \meter \squared} \\
		$r_R$ & 0.318 & \si{\meter} \\
		$m_M$ & 0.886 & \si{\kilogram} \\
		$I_S$ & 4.91e-3 & \si{\kilogram \meter \squared} \\
		$\rho$ & 0.02 & \si{\meter} \\
		$l$ & 0.09 & \si{\meter} \\  
		$d$ & 0.28 & \si{\meter} \\
		$b_R$ & 9.2e-3 & \si{\newton \per \meter \second} \\
		$b_C$ & 2.51e-2 & \si{\newton \per \meter \second} \\
		$b_S$ & 9.76e-3 & \si{\newton \per \meter \second} \\
		$u_{\max}$ & 4.27 & \si{\newton \meter}
	\end{tabular}
	\caption{Measured and identified parameters of the experimental single-rod brachiation robot.}\label{tab:parameters}
\end{table}
\begin{figure}[tb!]
    \centering
    \includegraphics[width=\linewidth]{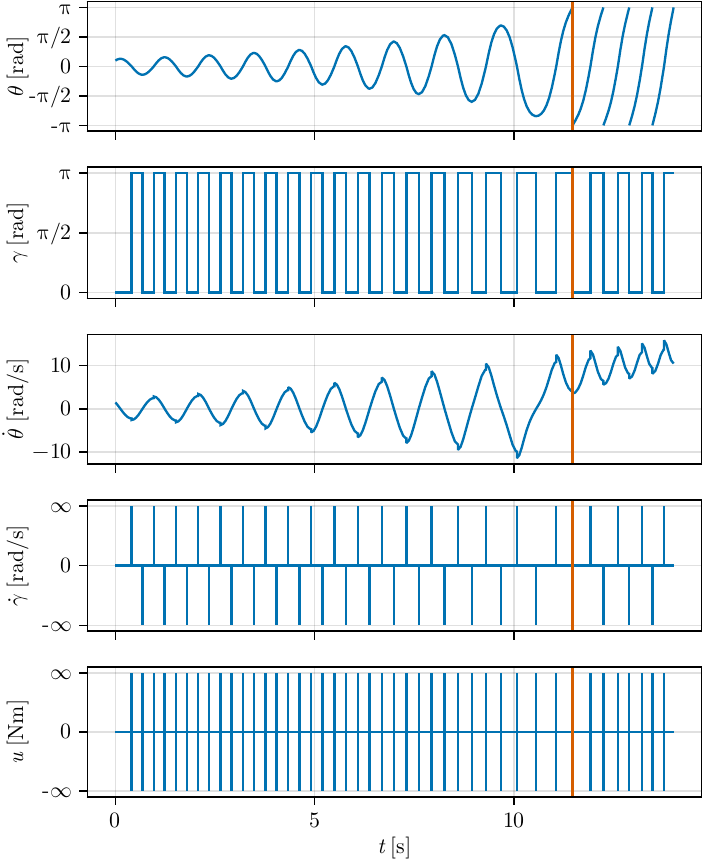}
    \caption{Evolution of the system's states and input during the execution of the limit case control policy in simulation. An orange vertical line denotes the instant at which the angle $\theta = \pi\,\si{\radian}$ was crossed.}\label{fig:simulation:lc:states}
\end{figure}
\begin{figure}[tb!]
    \centering
    \includegraphics[width=\linewidth]{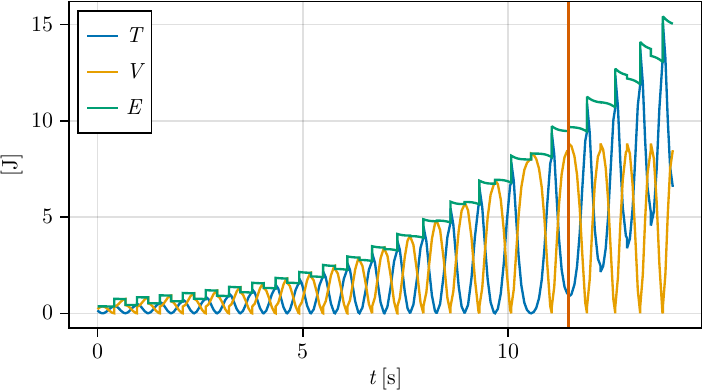}
    \caption{Evolution of the system's kinetic, potential, and total energy during the execution of the limit case control policy in simulation, corresponding to \cref{fig:simulation:lc:states}. The kinetic energy associated with the radial motion of the weight is omitted to maintain legibility as it would result in spikes reaching infinity.} %An orange vertical line denotes the instant at which the angle $\theta = \pi\,\si{\radian}$ was crossed.}h
    \label{fig:simulation:lc:energy}
\end{figure}

\subsection{Simulations of Limit Case Control}\label{sec:validation:lc}

First, the limit case policy is simulated, for which the model is supplemented by the jump map \cref{eqn:switch}, with \cref{eqn:jump_set:all}-\cref{eqn:jump_map}. The simulations have been performed in the Julia programming language, using the DifferentialEquations.jl suite of solvers, more precisely a 5th-order adaptive Runge-Kutta method with the absolute and relative tolerances of $10^{-7}$ and $10^{-5}$, respectively. To implement the policy, callbacks provided by the suite of solvers have been used to detect zero-crossings and consequently modify the system's state. The results of the simulation, starting from the initial state $x(0) = {[0.31 \ 0 \ 1.46 \ 0]}^\top$ (approximately matching the state of the system at the time of the controller's activation during the experiment later described in \cref{sec:validation:ex}) are shown in \cref{fig:simulation:lc:states}. As can be seen, despite the idealized unlimited control torque $u$, forming a set of weighted Dirac impulses to change the angle $\gamma$ step-wise between $0$ and $\pi$, the amplitude of $\theta$ grows relatively slowly. Notice that the limit angle $\theta=\pi\,\si{\radian}$ is crossed at $t=\SI{11.53}{\second}$ after $9$ swing periods. At this instant of time, the swinging stage is changed to the second rotating stage and the active jump set in \cref{eqn:jump_map} changes from \cref{eqn:jump_set:D1} and \cref{eqn:jump_set:D2} to \cref{eqn:jump_set:D1} and \cref{eqn:jump_set:D3}. After four revolutions, the simulation is terminated at $t=\SI{14.01}{\second}$ with a terminal angle of $\theta=9\pi\,\si{\radian}$. The corresponding energy balance is shown in \cref{fig:simulation:lc:energy}. In addition to kinetic $T$ \cref{eqn:T} and potential $V$ \cref{eqn:V} energy, the total energy $E = T+V$ is also visualized. As can be seen in \cref{fig:simulation:lc:energy}, during the swinging stage, both the kinetic and potential energy experience a large step-wise increase as the robot's body passes its stable equilibrium. This is followed by a gradual decrease due to damping in the system and a smaller step-wise decrease or increase in the potential energy as the mass is repositioned at the turning angle. This change in potential energy is negative if $|\theta| < \frac{\pi}{2}$ and positive if $|\theta| > \frac{\pi}{2}$. During the rotation phase, increases in both the kinetic and potential energy while passing the stable equilibrium are paired with a decrease in the kinetic energy and increase in the potential energy as the robot's body passes the unstable equilibrium. 
\begin{figure}[tb!]
    \centering
    \includegraphics[width=\linewidth]{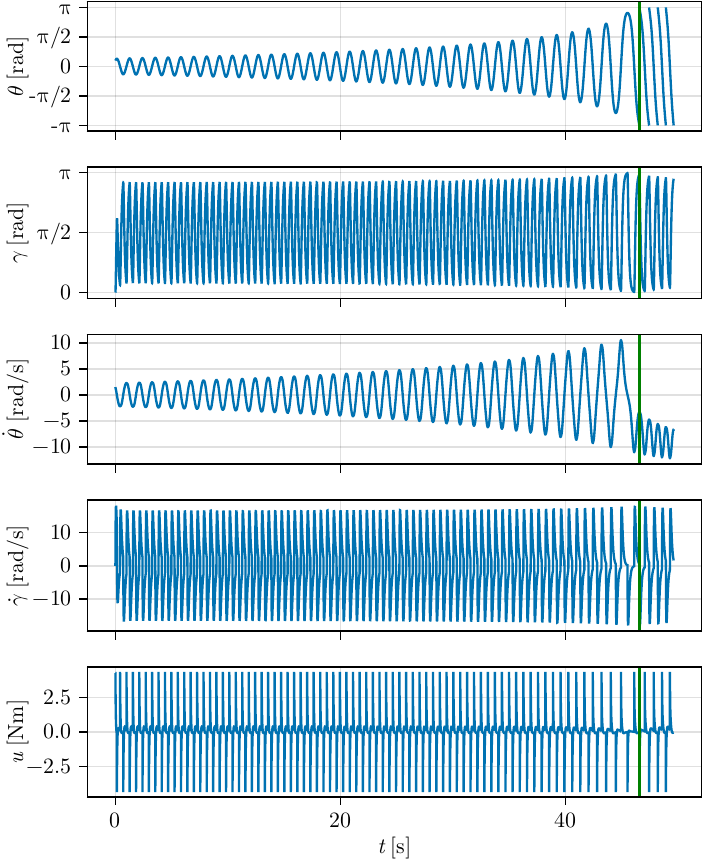}
    \caption{Evolution of the system's states and input during the execution of the continuous control policy in simulation. A green vertical line denotes the instant at which the angle $\theta = -\pi\,\si{\radian}$ was crossed.}\label{fig:simulation:io:states}
\end{figure}
\begin{figure}[tb!]
    \centering
    \includegraphics[width=\linewidth]{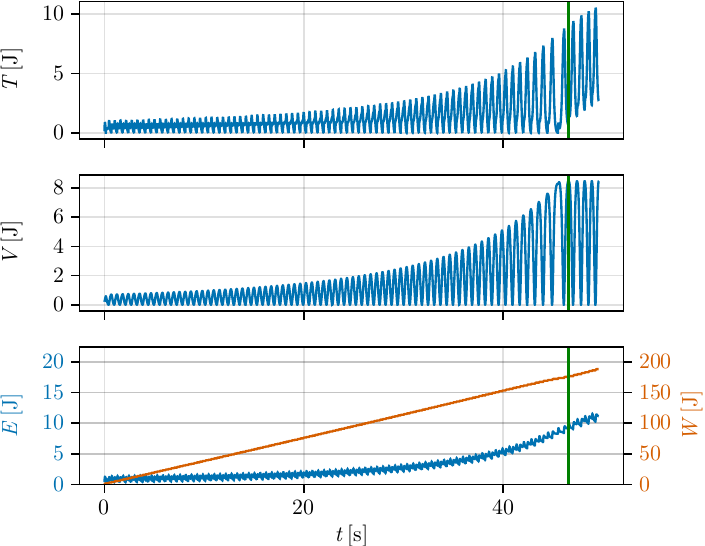}
    \caption{Evolution of the system's kinetic, potential, total, and input energy during the execution of the continuous control policy in simulation, corresponding to \cref{fig:simulation:io:states}}%A green vertical line denotes the instant at which the angle $\theta = -\pi\,\si{\radian}$ was crossed.}
\label{fig:simulation:io:energy}
\end{figure}

\subsection{Simulations of Continuous Control}\label{sec:validation:pd}

After analyzing the limit case policy showing the best achievable results for the given construction of the system, let us parameterize the continuous control policy and validate it in simulation. Taking into account the maximum available torque $u_{\max}$ provided in \cref{tab:parameters} and setting $\zeta=1$, the frequency of the target second-order dynamics \cref{eqn:2ndorder} is assigned to $\omega=\SI{17.14}{\per \second}$, based on \cref{eqn:omega}. Considering the setpoint generator \cref{eqn:gamma_d} and the control algorithm \cref{eqn:signal}-\cref{eqn:input}, the simulation results starting from the identical initial state as in the limit case, i.e.\ with $x(0) = {[0.31 \ 0 \ 1.46 \ 0]}^\top$, are shown in \cref{fig:simulation:io:states}. For this practical case, the limit angle $\theta=-\pi\,\si{\radian}$ is crossed at $t=\SI{46.55}{\second}$ after $37.5$ swing periods, that is, after $32.55$ more compared to the limit case. After four revolutions, the simulation terminates at $t=\SI{49.59}{\second}$ with a terminal angle of $\theta=-9\pi\,\si{\radian}$. 

As expected, when comparing the two simulations, we may observe that the terminal angle was reached earlier in the limit-case simulation. Notably, in both simulations there are sharp increases in the angular velocity of the robot when the weight is repositioned while passing the stable equilibrium. This coincides with the explanation that the motion is amplified by the Coriolis force when utilizing the continuous controller and also supports the claim that angular momentum of the robot must be preserved during the state jumps of the limit-case. These aspects also project to the energy evolution in \cref{fig:simulation:io:energy}. Here we may notice large spikes in the kinetic energy instead of step-wise changes which can be observed in \cref{fig:simulation:lc:energy}. These are present because the radial motion of the mass, which accelerates and decelerates as it is repositioned, is also included in the kinetic energy of the system. Otherwise, the evolution of the system's energy is similar to that in \cref{sec:validation:lc}.

Regarding the total accumulated energy, it increased from $E(0)=\SI{0.35}{J}$ to $E(49.5)=\SI{11.15}{J}$, i.e. with $\Delta E=\SI{10.8}{J}$. Energy added to the system $W$, determined as the integral of power supplied by the actuator
\begin{equation}
    P(t) = \begin{cases}
        \dot{\gamma} \, u ,& \text{if } \dot{\gamma} \, u > 0 \\
        0 ,& \text{else }
    \end{cases},
\end{equation}
increases almost linearly from $W(0)=\SI{0}{J}$ to $W(49.5)=\SI{188.65}{J}$. Thus, the energy efficiency is relatively small, corresponding to $5.7\%$.

\begin{figure}[tb!]
    \centering
    \includegraphics[width=\linewidth]{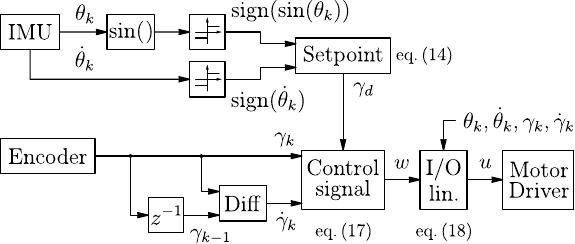}
    \caption{Block diagram of the control scheme implemented in the {MCU}.}
    \label{fig:control_scheme}
\end{figure}
\begin{figure*}[tb!]
    \centering
    \includegraphics[width=\linewidth]{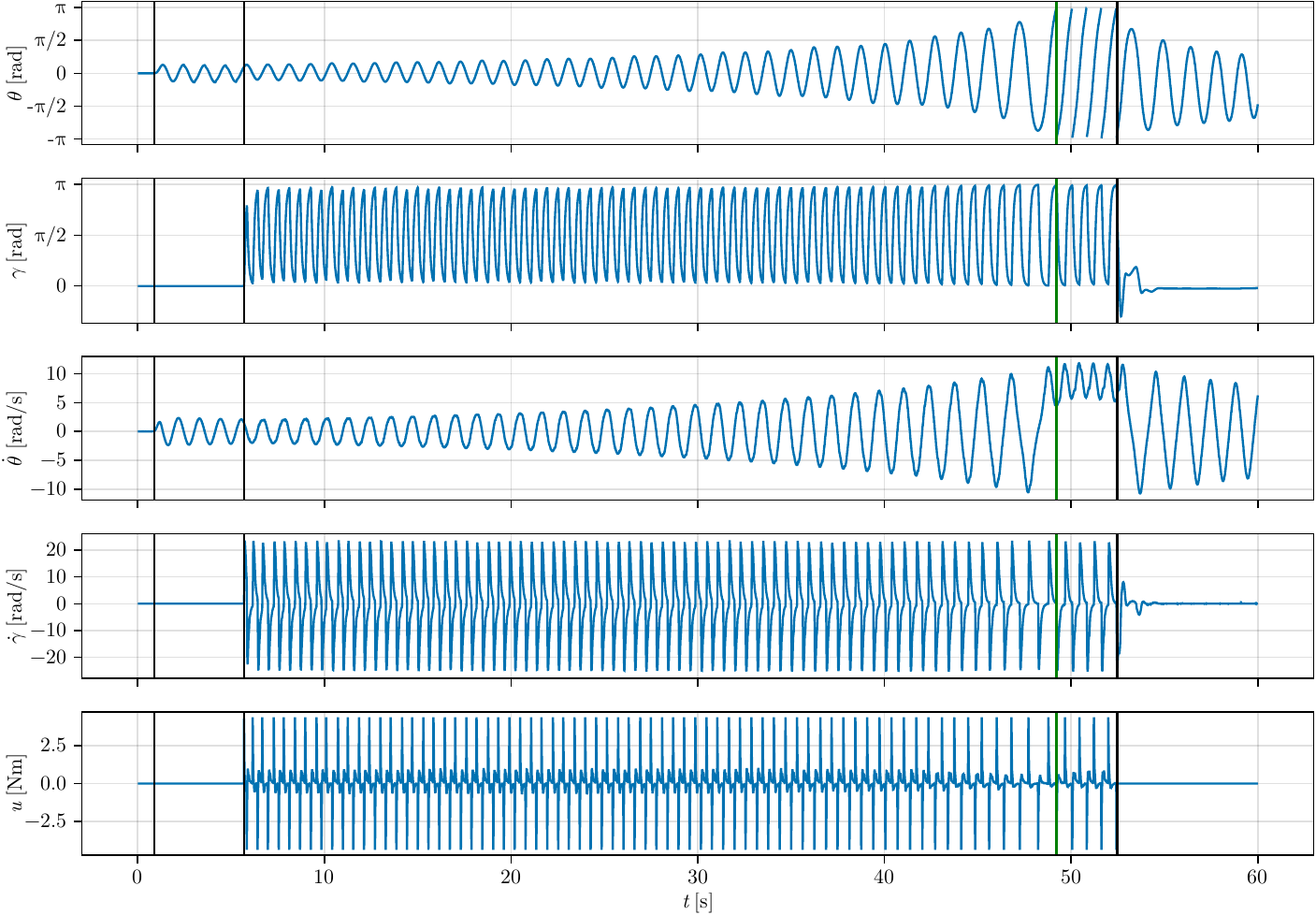}
    \caption{Evolution of the system's states and input during the experimental validation of the continuous control policy. Black vertical lines, in sequence, denote the initial perturbation, controller enable, and controller disable. The green vertical line denotes the instant at which the angle $\theta = \pi\,\si{\radian}$ was crossed.}\label{fig:experiment:states}
\end{figure*}
\begin{figure*}[t]
    \centering
\includegraphics[width=\linewidth]{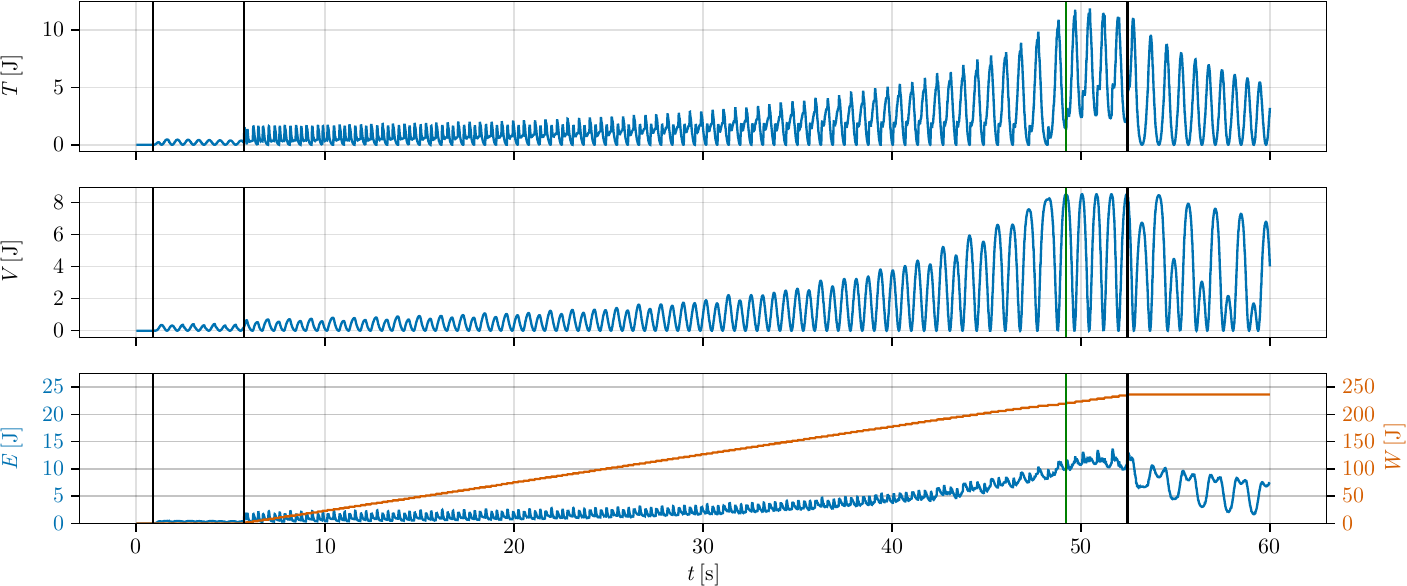}
    \caption{Evolution of the system's kinetic, potential, total, and input energy during the experimental validation of the continuous control policy, corresponding to \cref{fig:experiment:states}.} %Black vertical lines, in sequence, denote the initial perturbation, controller enable, and controller disable. The green vertical line denotes the instant at which the angle $\theta = \pi\,\si{\radian}$ was crossed.}
    \label{fig:experiment:energy}
\end{figure*}

\subsection{Experiments of Continuous Control}\label{sec:validation:ex}
Finally, the experimental validation of the continuous control policy parameterized above, i.e., with $\zeta=1$ and $\omega=\SI{17.14}{\per \second}$, is performed with an implementation of the algorithm illustrated in \cref{fig:control_scheme}. Angle $\theta$ and angular velocity $\dot{\theta}$ were provided by the {IMU}, while an encoder, mounted on the axle of the motor, supplied $\gamma$ from which $\dot{\gamma}$ was derived using finite differencing. 

The results of the experiment are shown in \cref{fig:experiment:states} and \cref{fig:experiment:energy}; see also the experiment video\footnote{\href{https://control.fs.cvut.cz/en/aclab/experiments/brachbotexp}{https://control.fs.cvut.cz/en/aclab/experiments/brachbotexp}}. The experiment starts with the system at the stable equilibrium, from which it is manually perturbed at $t=\SI{0.9}{\second}$. The controller is then enabled at $t=\SI{5.69}{\second}$, when $x(5.69) = {[0.31 \ 0.012 \ 1.46 \ 1.29]}^\top$ and $E(5.69) = \SI{0.35}{\joule}$, increasing the magnitude of oscillations until the angle of $\theta = \pi\,\si{\radian}$ is reached at $t=\SI{49.21}{\second}$. Consequently, four additional revolutions are performed. The angle $\theta = 9\pi\,\si{\radian}$ is reached at $t=\SI{52.42}{\second}$ with the total energy in the system $E(52.42) = \SI{10.7}{\joule}$, after which the controller is disabled at $t=\SI{52.47}{\second}$. The energy added to the system during the control period is $W = \SI{236}{\joule}$, from which we may derive the overall energy efficiency of $4.5\%$. 

When comparing the experimental (\cref{fig:experiment:states}) and simulation results (\cref{fig:simulation:io:states}), we can conclude that the fourth revolution was completed approximately $6\%$ faster in the experiment, but with a lower energy efficiency. Both can be attributed to an inevitable mismatch between the model and the experimental setup, which we regard as acceptable based on the results.

\section{Conclusions}\label{sec:conclusions}
In the paper, we have presented a general mechatronic concept and thorough validation of a single-rod brachiation robot in its pre-jump phases. The minimal construction towards brachiation is composed of a rod, fixed gripper mechanisms to hold a bar, and a crank-slide mechanism that repositions the robot's center of mass with the aim of amplifying the swing motion or enhancing the angular velocity during rotation. After the construction and hardware ({HW}) aspects of the mechatronic setup were addressed, the emphasis was placed on developing control strategies to pump energy into the system by repositioning the center of mass of the single-rod robot. The best results were obtained in simulation using a bang-bang limit-case control strategy, which, however, cannot be implemented on a physical device with limited actuation. The subsequently proposed continuous control strategy takes into account the limited torque of the servo and the kinematics of the crank-slide mechanism. Through simulations, it was demonstrated that the continuous control strategy is well applicable. The continuous control strategy is based on input-output linearization which turns the nonlinear crank-slide mechanism dynamics into linear second-order dynamics, which are parameterized based on the maximum torque of the motor. The final and main result presented is the experimental validation of the proposed continuous control strategy. The experiments clearly demonstrate the viability of the proposed minimal concept toward brachiation.

In follow-up research, the subsequent jump stage of the brachiation will be targeted. For that, the mechatronic concept of the robot needs to be further optimized. This will mainly include completely transitioning to wireless connectivity and will naturally be accompanied by an increase in weight as a result of the presence of a battery and additional HW on the rod. 
Along with the necessary redesign of the robot's construction, the path length, weight, and actuation power for the moving mass need to be optimized to achieve higher efficiency in transforming the energy from the actuator to robot motion. To improve the performance of the control algorithm, it can be supplemented by a more advanced state estimation approach or a model that also considers dry friction. An enhanced attention also needs to be paid to trajectory planning for the jump stage, providing the state at which the gripper mechanism at the current bar should be released in order to hit the target bar in desired position. The trajectory planning will naturally require a more powerful control system HW and implementation of the higher level predictive control layer. Another challenge is to compensate for the effect of disturbances within the overall control system of the brachiation robot.
\bibliographystyle{IEEEtran}
\bibliography{References,WCbib}
\def\biospace{-0.3in}

\begin{IEEEbiography}[{\includegraphics[width=1in,clip,keepaspectratio]{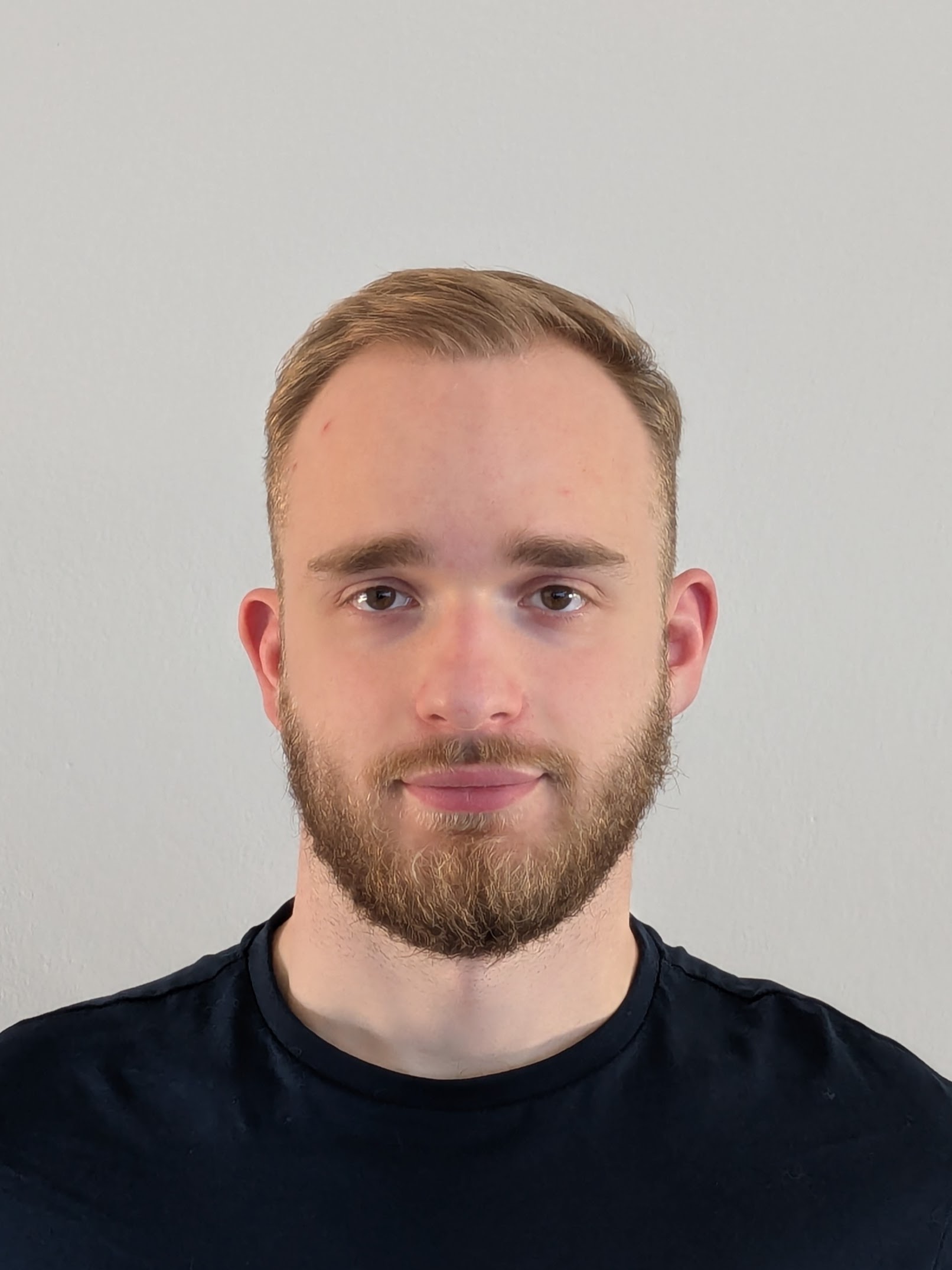}}]{Juraj Lieskovsk\'{y}}  
recieved his M.Sc.\ degree in Mechatronics from the Faculty of Mechanical Engineering (FME), Czech Technical University (CTU) in Prague, Czech Republic, in 2022. Since then, he has been a PhD student in Machine and Process Control at the Dept.\ of Instrumentation and Control Eng., FME-CTU.\@ His research interests include optimal control and state estimation, including model predictive control and moving horizon estimation, with a focus on nonlinear systems.
\end{IEEEbiography}
\vspace{\biospace}
\begin{IEEEbiography}[{\includegraphics[width=1in,clip,keepaspectratio]{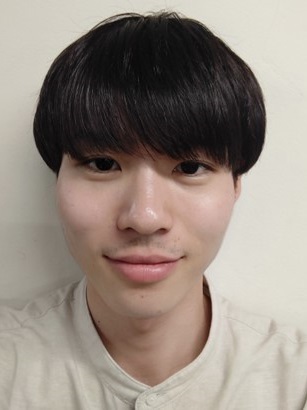}}]{Hijiri Akahane} recieved his M.Eng.\ degree from the Department of Mechanical Systems Engineering, Tokyo University of Agriculture and Technology, Japan, in 2024. His research interests include machine design, mechatronics, and bio-mimetic robotics.
\end{IEEEbiography}
\vspace{\biospace}
\begin{IEEEbiography}[{\includegraphics[width=1in,clip,keepaspectratio]{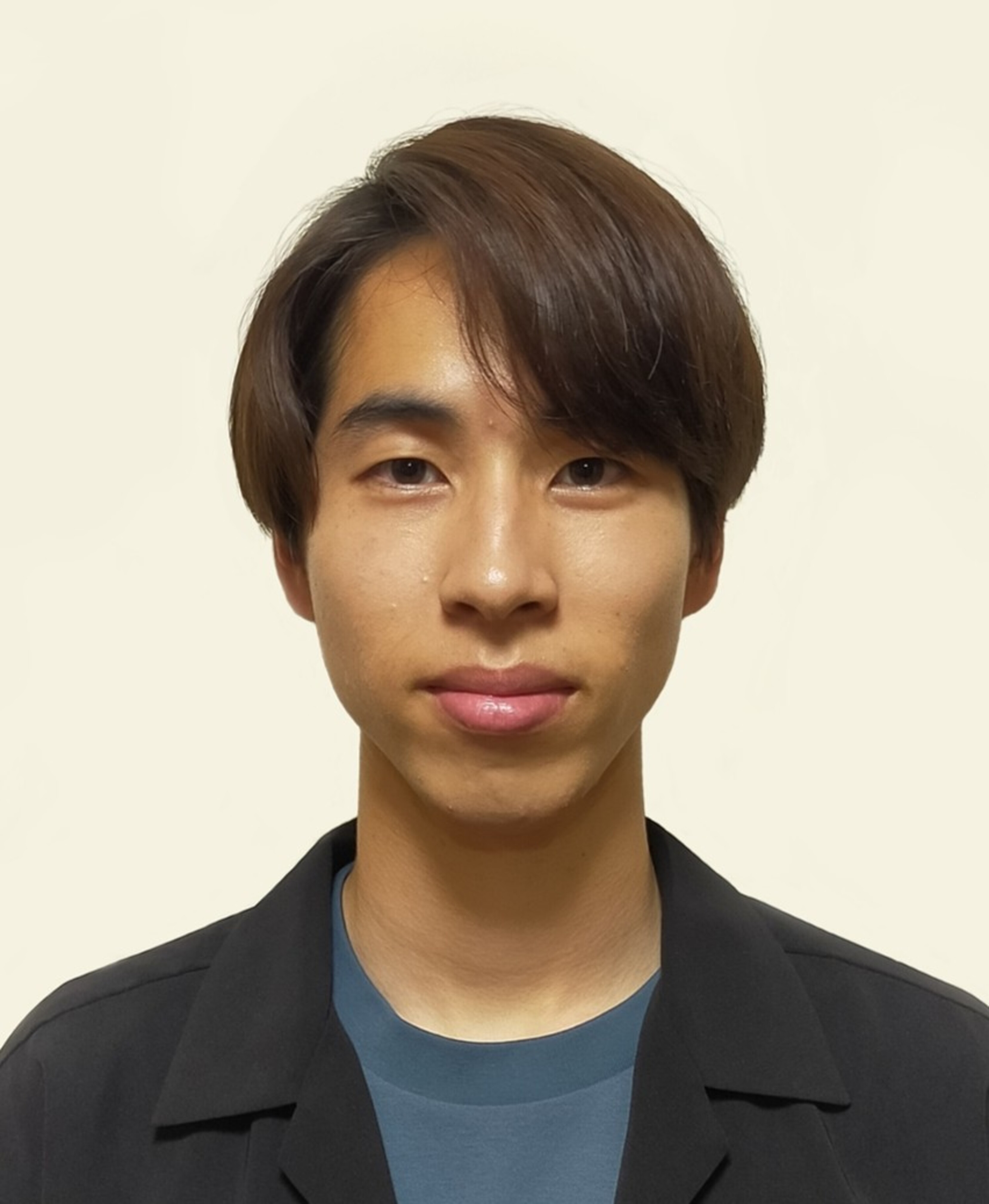}}]{Aoto Osawa} has been bachelor student in the Department of Mechanical Systems Engineering, Tokyo University of Agriculture and Technology, Japan, since 2021. His research interests include mechatronics, bio-mimetic robotics, and trajectory optimization.
\end{IEEEbiography}
\vspace{\biospace}
\begin{IEEEbiography}[{\includegraphics[width=1in,clip,keepaspectratio]{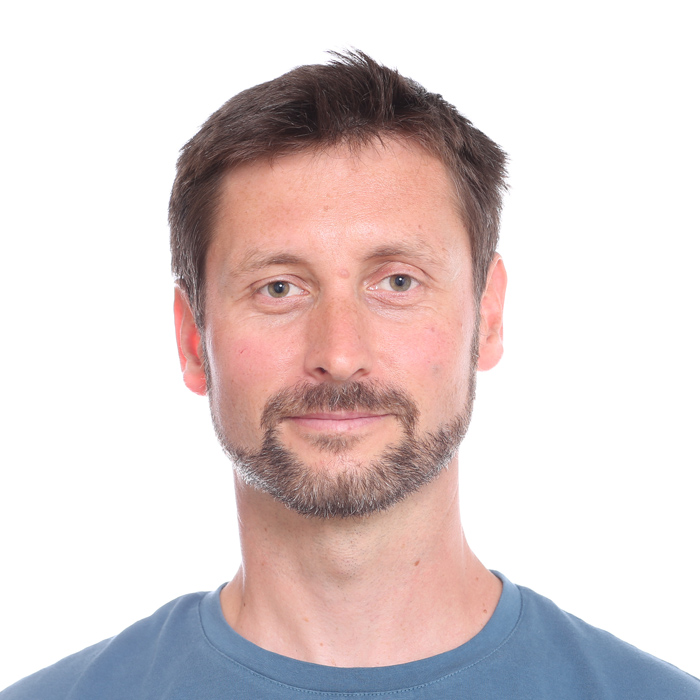}}]{Jaroslav Bušek}
received the B.Sc.\ in Information and Automation Technology (2009), M.Sc.\ in Instrumentation and Control Engineering (2011) and Ph.D. in Technical Cybernetics (2019), all from the Faculty of Mechanical Engineering (FME), Czech Technical University in Prague (CTU), Czechia. Since 2014, he has been with the Dept.\ of Instrumentation and Control Eng., FME – CTU, and since 2018 also with the Czech Institute of Informatics, Robotics and Cybernetics, CIIRC – CTU.\@ His research interests include mathematical modeling, practical aspects of time-delay system control, vibration suppression, and flexible mode compensation at smart mechanical structures.
\end{IEEEbiography}
\vspace{\biospace}
\begin{IEEEbiography}[{\includegraphics[width=1in,clip,keepaspectratio]{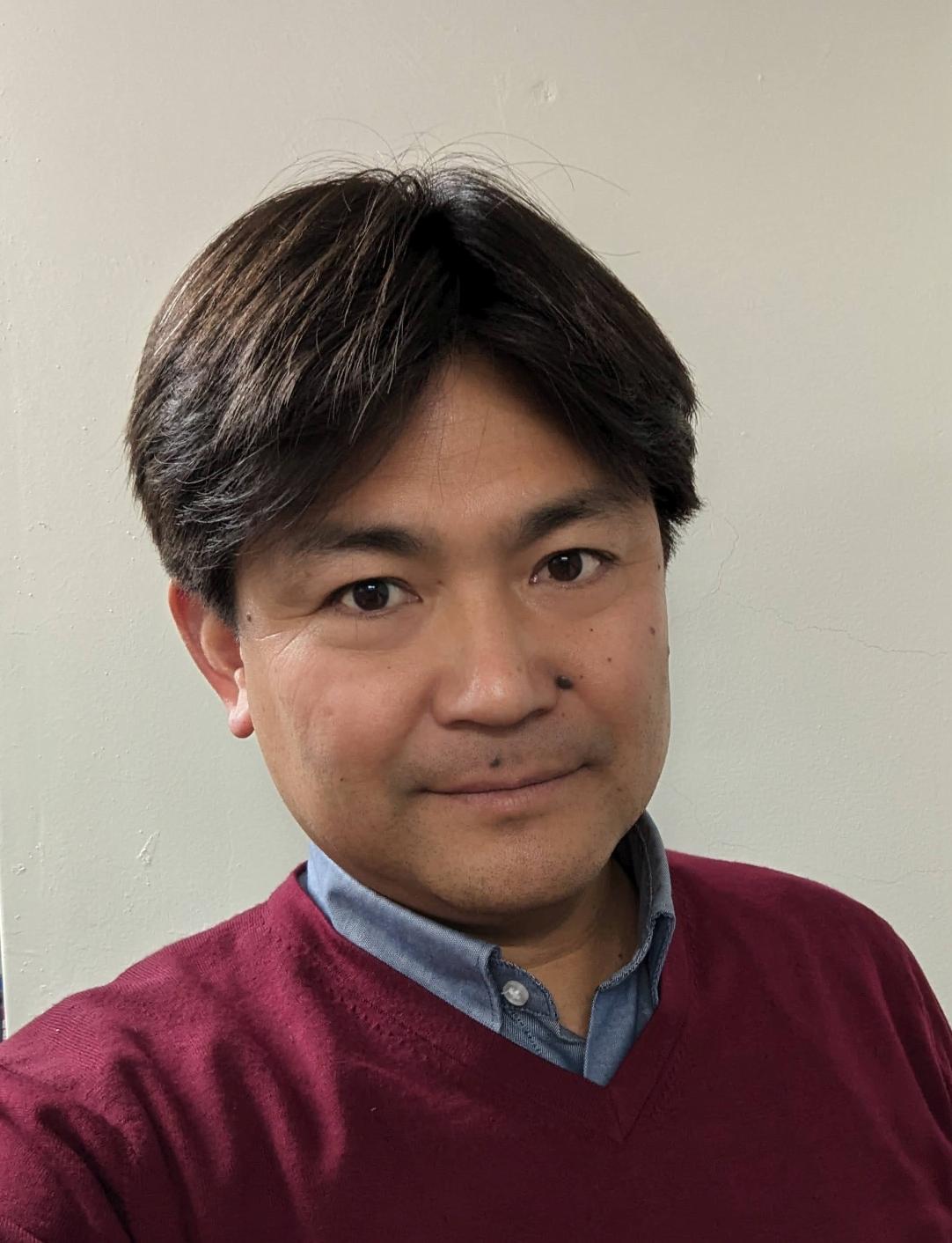}}]{Ikuo Mizuuchi} received the B.Eng.\ in Mechanical Eng.\ from Waseda University in 1995, and the M.Eng.\ and Ph.D. in Mechano-Informatics from the University of Tokyo in 1998 and 2002, respectively.  He was a Research Fellow (2000–2001) of the Japan Society for the Promotion of Science and a Project Assistant Professor with the University of Tokyo (2002–2006) and then became a Senior Assistant Professor with the Dept of Mechano-Informatics.  He was a Member of JSK Robotics Laboratory and developed several musculoskeletal humanoids, including Kenta, Kotaro, and Kojiro. In 2009, he started Mizuuchi Lab.\ at Tokyo University of Agriculture and Technology, as an Associate Professor, and became a Professor there in 2020.  His research interests include humanoids, home robots, agricultural robots, and human-inspired systems.  Dr.\ Mizuuchi received the Best Paper Award of Advanced Robotics and the Best Conference Paper Award Finalist at the Int. Conf.\ on Robotics and Automation.
\end{IEEEbiography}
\vspace{\biospace}
\begin{IEEEbiography}[{\includegraphics[width=1in,clip,keepaspectratio]{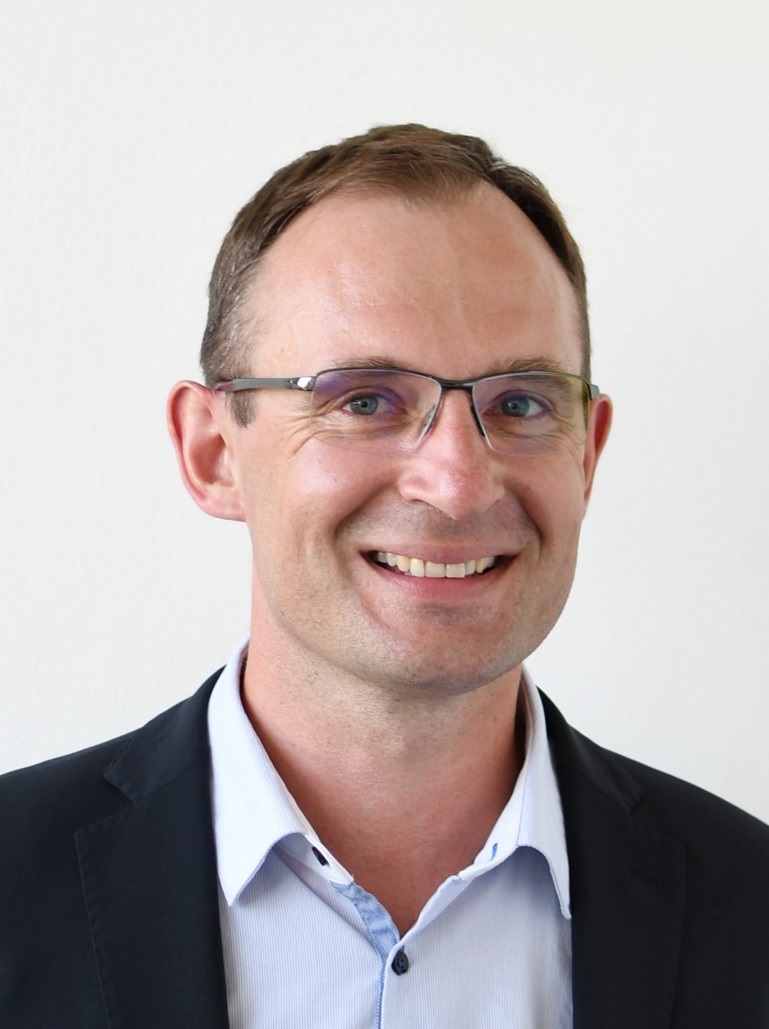}}]{Tom\'{a}\v{s} Vyhl\'{\i}dal} graduated in Automatic Control and Engineering Informatics in 1998 and received Ph.D. in Control and Systems Engineering in 2003, both from the Faculty of Mechanical Engineering (FME), Czech Technical University in Prague (CTU). In 2012, he became a professor at CTU in the subject of Control and Systems Engineering. Since 2000, he has been with the Dept. of Instrumentation and Control Eng. (dept. head since 2019), FME – CTU, and since 2015 also with the Czech Institute of Informatics, Robotics and Cybernetics, CIIRC - CTU. His research interests include analysis and control design of time-delay systems, mathematical modelling and applied control theory. He has been a member of the IFAC Technical Committee for Linear Control Systems since 2013, vice-chair for industry 2017-2023.
\end{IEEEbiography}

\end{document}